\documentclass[a4paper,10pt]{article}
\usepackage[utf8]{inputenc}

\usepackage{url,hyperref}
\usepackage{graphicx}  
\usepackage{subcaption} 
\usepackage{geometry}
\title{SocNav1: A Dataset to Benchmark and \\ Learn Social Navigation Conventions\footnote{\normalsize Submitted to \textit{Data}, MDPI, on January 11, 2020.}}
\author{Luis J. Manso,\footnote{Computer Science Department, School of Engineering and Applied Science, Aston~University, Birmingham, United~Kingdom. {\textbf{l.manso@aston.ac.uk}}} ~Pedro N\'u\~nez,\footnote{Robotics and Artificial Vision Laboratory - RoboLab, C\'aceres School of Technology, Universidad de {Extremadura}, C\'aceres, Extremadura.}  ~Luis V. Calderita,$^{\ddag}$ \\ Diego R. Faria$^{\dagger}$ and Pilar Bachiller\,$^{\ddag}$}

\date{January 11, 2020 }

\usepackage[labelformat=simple]{subcaption} 

\DeclareCaptionLabelFormat{subcaptionlabel}{\normalfont(\textbf{#2}\normalfont)}
\begin{document}
\maketitle

\abstract{Datasets are essential to the development and~evaluation of machine learning and~artificial intelligence algorithms. As new tasks are addressed, new datasets are required. Training algorithms for human-aware navigation is an example of this need. Different factors make designing and~gathering data for human-aware navigation datasets challenging. Firstly, the problem itself is subjective, different dataset contributors will very frequently disagree to some extent on their labels. Secondly, the number of variables to consider is undetermined culture-dependent. This paper presents SocNav1, a dataset for social navigation conventions. SocNav1 aims at evaluating the robots' ability to assess the level of discomfort that their presence might generate among humans. The~9280 samples in SocNav1 seem to be enough for machine learning purposes given the relatively small size of the data structures describing the scenarios. Furthermore, SocNav1 is particularly well-suited to be used to benchmark non-Euclidean machine learning algorithms such as graph neural networks. This paper describes the proposed dataset and~the method employed to gather the data. To provide a~further understanding of the nature of the dataset, an analysis and validation of the collected data are also presented.}

\section{Summary}\label{summary}
Adapting to social conventions is an unavoidable requirement for the acceptance of assistive and~social robots.
While the scientific community broadly accepts that assistive robots and~social robot companions are unlikely to have widespread use in the near future, their presence in health-care and~other medium-sized institutions is becoming a reality.
These robots will have a beneficial impact in industry (see~\cite{rajnai2017labor, pham2018impact}) and~other fields such as health care (see~\cite{fosch2018did, garcia2018challenges}).
The growing number of research contributions to social navigation is also indicative of the importance of the topic.
To foster the future prevalence of these robots, they must be useful, but also socially accepted.
As first proposed by~\cite{Trinh2015} and~later by~\cite{vega2019socially} and~\cite{Cruz-maya2019}, robots should navigate politely, actively asking for permission or collaboration when necessary.
The first step to be able to actively ask for collaboration or permission is to estimate whether the robot would make people feel uncomfortable otherwise, and that is precisely the goal of algorithms estimating social navigation compliance.
Some approaches provide analytic models, whereas others use machine learning techniques such as neural networks (see~\cite{perez2018learning}).
Regardless of the approach followed, modelling social conventions is very challenging.
Firstly, because the problem itself is subjective.
Secondly, because of the variables involved, whose number and~weight is undetermined and~changing.
This paper presents and~describes SocNav1, a dataset for social navigation conventions.
The aims of SocNav1 are two-fold: (a) enabling comparison of the algorithms that robots use to assess the convenience of their presence in a particular position when navigating; (b) providing a sufficient amount of data so that modern machine learning algorithms such as deep neural networks can~be~used.
Because of the structured nature of the data, SocNav1 is particularly well-suited to be used to benchmark non-Euclidean machine learning algorithms such as graph neural networks (see~\cite{Battaglia2018,manso2019graph}).
The dataset has been made available in a public repository\footnote{\url{https://github.com/ljmanso/SocNav1}}.

\par
There are many different factors that influence robot social acceptance (\cite{Rios-Martinez2015}), including visual appearance, interaction skills and~an appropriate management of the interaction spaces.
The~study~of~how humans manage their interaction distances with other people is called proxemics~(\cite{hall1969hidden}).
Multiple~social navigation approaches build on the idea of proxemics to improve robots' social acceptability in~navigation (e.g.,~\cite{hansen2009adaptive, Ramon-Vigo2014}).
However, as pointed out by \cite{Rios-Martinez2015}, there are other factors that should~be~taken into account to avoid disturbing humans, such as human interaction groups, information process spaces or affordance and~activity spaces.
Some of these concepts have been incorporated in studies where an analytic solution is provided (e.g.,~\cite{Borkowski2010, vega2019socially}), whereas others follow a machine learning approach (e.g.,~\cite{hansen2009adaptive, Ramon-Vigo2014}).
Independently of the nature of its implementation, the~importance of social navigation makes key having appropriate datasets, not only for benchmarking but also for learning purposes.  

\par
Several public datasets have been used in social navigation. 
In~\cite{Luber2012}, authors use the Edinburgh Informatics Forum Pedestrian Database (EIPD) to make a robot learn the behaviour of pedestrians.
Another interesting dataset is the one used in~\cite{Papadakis2013}, which contains recorded action sequences that correspond to social interactions.
The authors use it in a social mapping approach.
In~\cite{Fisher2004}, a dataset for public space surveillance task was also made public.
It consists of 28 video sequences of 6 different scenarios.
Two data sets are also described in~\cite{Pellegrini2009} for tracking multiple people.
The dataset was acquired from birds-eye and~manually annotated.  

\par
To the best of our knowledge, the social navigation datasets available in the literature provide data to benchmark and/or learn route estimators based on the behaviour of humans.
The first motivation to~generate a new dataset is that, especially while the technology readiness level is not high enough, the~behaviour that humans expect from robots might be different from the one expected from fellow humans.
Generally, humans would expect robots to keep a safer distance in comparison to other humans.
Among the possible causes of this phenomena, we can highlight the noise made when robots move, and~the apparent unpredictability of their behaviour in comparison to that of humans.
The~second motivation of the dataset is that SocNav1 aims at evaluating the robots' ability to assess the level of~discomfort that their presence might generate among humans.
This ability would be used~by~robot navigation systems to estimate path costs, but SocNav1 does not directly deal with path costs.

\par
The remainder of the paper is as follows. Section~\ref{data_description} describes the dataset. Section~\ref{methods} describes the~methods used to collect data. Section~\ref{analysis} provides an analysis and validation of the data collected. A discussion on the advantages and~limitations of SocNav1 is provided in Section~\ref{discussion}.

\section{Data Description}\label{data_description}
\par
The dataset is composed of four JSON files: three files for training, development and~testing, and~a fourth file for training with data augmentation.
The percentage of samples for the training, development and~testing datasets were 88\%, 6\% and~6\% respectively.
The split was made to ensure a fair comparison of the algorithms using the dataset, especially given the high number of samples in comparison to the size of each scenario description.
In addition, data augmentation was carried out by mirroring the scene over the frontal axis, assuming that mirrored scenarios should have the same labels.
The samples were shuffled before splitting the dataset into train/dev/set.
The augmented dataset was also shuffled after the augmentation process.
The main files in the dataset are located in~the data subdirectory:
\begin{itemize}
\item socnav\_training.json: training dataset. No data augmentation. It contains 8168 labels/scenarios.
\item socnav\_training\_dup.json: training dataset with data augmentation. It contains 16336 labels/scenarios.
\item socnav\_dev.json: development dataset. It contains 556 labels/scenarios.
\item socnav\_test.json: testing dataset. It contains 556 labels/scenarios.
\end{itemize}

\par
Each line in these files contains a description of a labelled scenario, which is a representation of a scene at a given time during robot navigation.
They contain the following elements:
\begin{itemize}
    \item identifier: a string that identifies the scenario. Several instances of the same labelled scenario might exist.
    \item robot: it is a dictionary containing the identifier of the robot in the scenario.
    \item room: a list of points defining the wall polyline that delimits the room.
    \item humans: a list of humans. Each human is implemented as a dictionary with the following keys:
        id (identifying the human in the scenario),
        xPos and yPos (they are the center of the human and represent its location expressed in centimetres),
        orientation (expressed in degrees).
        Humans are assumed to be 40~cm wide, and 20~cm from chest to back.
    \item objects: a list of objects. Each object is implemented as a dictionary with the following keys:
        id (identifying the object in the scenario),
        xPos and yPos (the location of the object, expressed in centimetres),
        orientation (expressed in degrees).
        Objects are assumed to be $40\times40$~cm.
    \item links: a list of interaction tuples, where the first element of the tuple is a human who is interacting with the second element in the tuple, which can be an object or another human. 
    \item score: the score assigned to the robot in the scenario. From 0 to 100.
\end{itemize}

\par
The dataset was generated using two sets of possible scenarios.
The second set was created to increase the number of different scenarios. 
Both of them were randomly generated and the two include a wide variety of potential situations.
Using the first subset, composed of 2500 scenarios, three subjects generated a total of 5522 labels for the scenarios.
These scenarios were labelled multiple times with some level of disagreement between humans, as the nature of the problem is subjective.
Using the second subset, composed of 10000 scenarios, nine subjects generated a total of 3758 labelled scenarios with a low number of duplicates.
As a result, 12 subjects generated 9280 labels for the scenarios provided.
The age of the participants was between 22 to 45 years, 30\% of whom were women and the rest men.
They are native middle class residents of Spain.
Three of the subjects were researchers involved in the project, the rest were computer science students with no domain knowledge beyond the instructions they were given.
A total of 5735 different scenarios were used, 2761 were labelled once, 2406 were labelled twice and~568 were labelled three or more times.
When the dataset was designed, labelling scenarios multiple times was considered beneficial to evaluate to what extent humans agree on the labelling (see~Section~\ref{analysis}).
The whole data collection process took place between April 13th and April 27th, 2019.

Besides the data sub-directory, the repository has two other sub-directories: {raw\_data}, which contains the data collected by each of the 12 subjects, and~unlabelled, where the two subsets of scenarios used can be found (following the same file format and~a score of 0 for all the scenarios). All angles are~expressed in degrees, distances are expressed in centimetres.

\section{Methods}\label{methods}
\begin{figure}
\centering
    \subcaptionbox{Scenario A: a single person.\label{fig:scenario_a}}{\includegraphics[width=.45\textwidth]{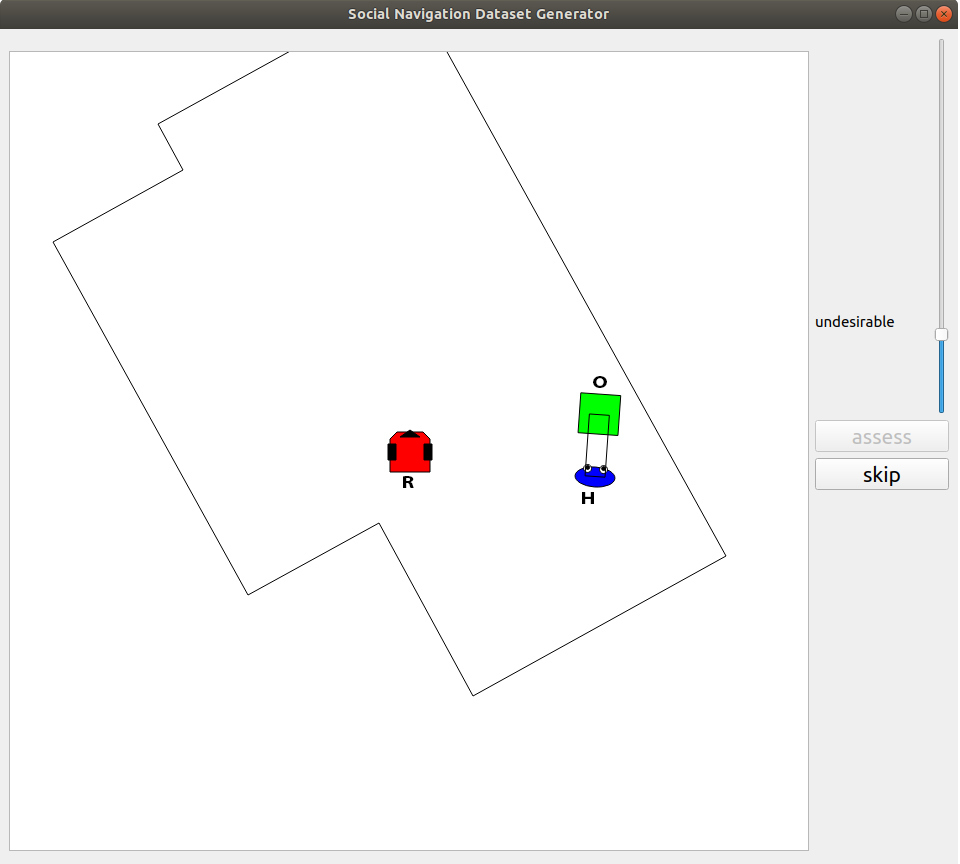}}
    \hfill
    \subcaptionbox{Scenario B: 2 people.\label{fig:scenario_b}}{\includegraphics[width=.45\textwidth]{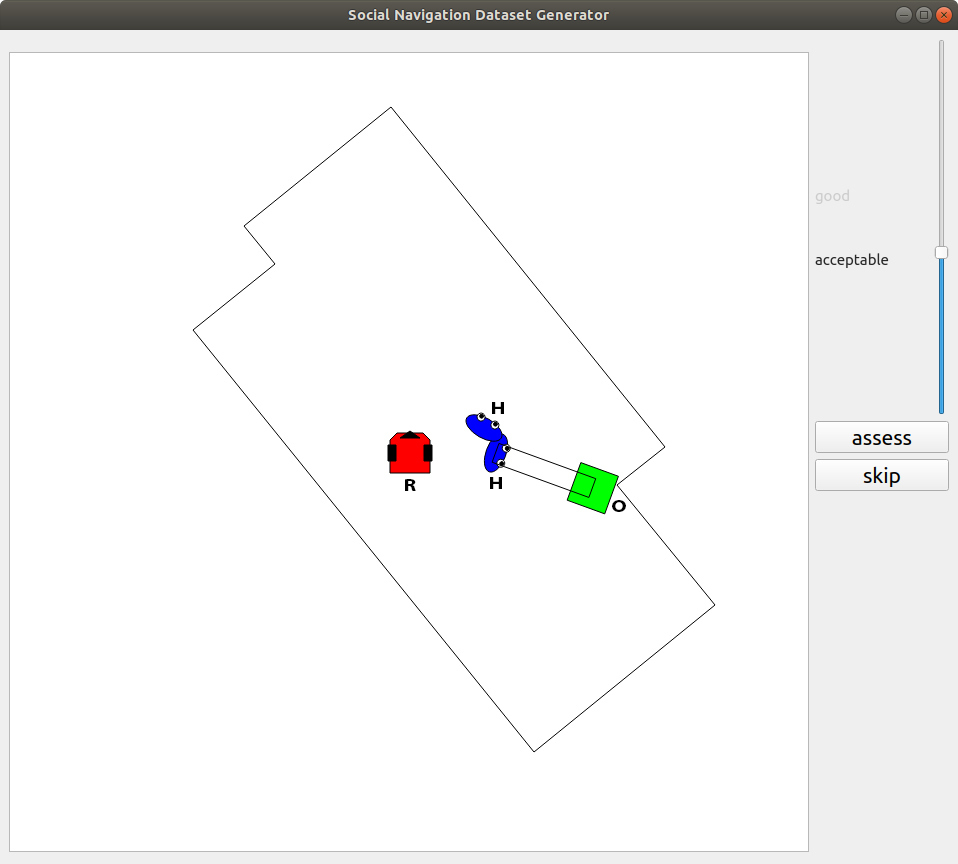}}
    \par\bigskip
    \subcaptionbox{Scenario C: 6 people.\label{fig:scenario_c}}{\includegraphics[width=.45\textwidth]{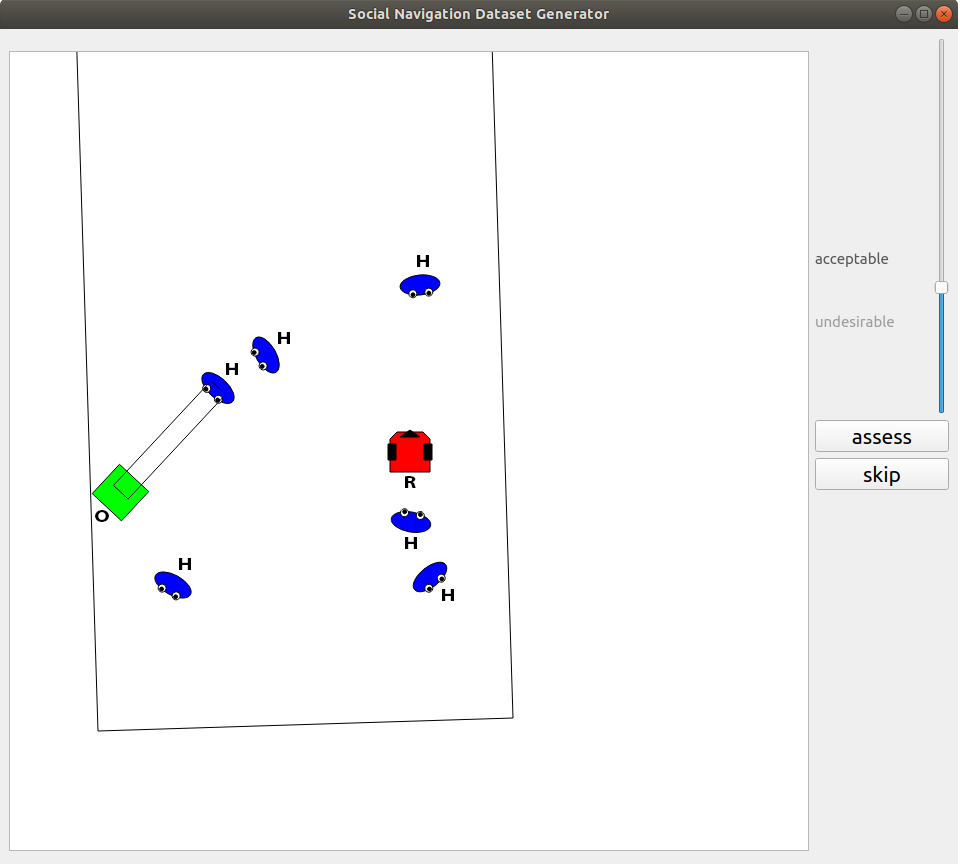}}
    \hfill
    \subcaptionbox{Scenario D: 17 people.\label{fig:scenario_d}}{\includegraphics[width=.45\textwidth]{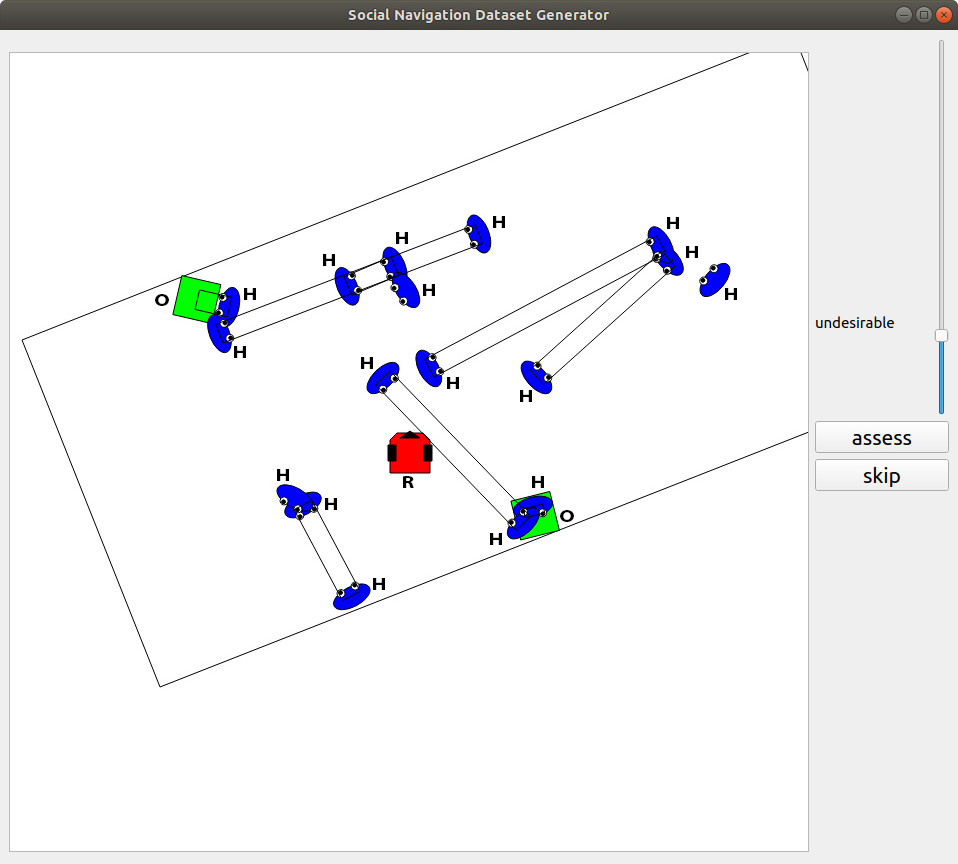}}
    \par\bigskip
    \caption{Screen shots of the application used to gather the data. The robot is depicted in red and labelled with R, the humans in blue and labelled with H and the objects in green and labelled with O. Black parallel lines indicate interaction between a human and an object or another human.}\label{fig:sndg}
\end{figure}

In order to acquire data at a feasible cost and~gather robot-specific information (i.e., not imitating the behaviour of humans), it was decided to develop an ad hoc application depicting the scenarios that humans had to manually assess (see Figure~\ref{fig:sndg}).

\par
The interface of the tool has two main areas.
The canvas on the left-hand side is used to depict the scenarios where subjects were asked to assess the robot's behaviour in terms of the disturbance caused to humans.
On the right hand side, users have a slider which value goes from 0 (unacceptable) to 100 (perfect). The intermediate labels are undesirable, acceptable, good and very good.
The interface smoothly transitions from one label to another using font transparency to make easier selecting intermediate values.
Also on the right-hand side, users can make use of two buttons, one to assess the current scene and generate a new one (button on top) and another one to avoid labelling the current scene in case they are unsure of how to label a particular scene (button on the bottom).

\par
The scenarios, rooms randomly generated under some restrictions to make them feasible, depict~$8\times8$ metres square areas where different elements can be found: the robot, walls, humans, objects and interaction indicators.
The generative process randomizes most of the elements in a scenario: the number and locations of the humans, the structure of the room, the number of objects and the interactions between humans and between humans and objects. This randomization favours the diversification of the potential situations represented by the different scenarios.
\par
The representation is robot-centric, so the robot ---in red and labelled with R--- is always in the centre of the canvas and~aligned with the axes.
There is always a room composed of, at least, four walls represented by black lines. 
Humans ---in blue and labelled with H--- and objects ---in green and labelled with O--- can be anywhere in the room.
They are only generated within the canvas, enforcing that even if the room is bigger than the canvas users will not miss any element.
Interactions are represented by parallel lines.
These might exist between humans ---for human-to-human interactions--- or between a human and an object ---to represent any kind of interaction with objects.
Interaction are composed of two elements, the first one is a human and the second one is an object or another human.
The positions of the two elements and the orientation of one of them are randomly generated.
The second element is assigned the opposite orientation of the first one.
The black lines representing interactions are drawn according to the positions of the two interacting elements.

\par
Despite some guidelines were provided, subjects were asked to feel free to express how they thought they would feel in the scenarios.
It must be noted that subjects are not asked about any particular human.
They should annotate the scenarios according to the disturbance caused by the robot to the humans in general.
The guidelines were the following:
\begin{itemize}
\item The closer the robot is to humans from their perspective, the more it disturbs.
\item A collision with a human should have a 0 score (unacceptable).
\item We want to consider, not only the personal spaces but also the spaces that humans need to~interact with other humans or objects. The closer the robot gets to the interaction space (human to human, or human to object) the lower the score ---up to a non-critical limit.
\item A collision with an interaction area should have a maximum score of 20 (undesirable). The interaction area is considered the zone needed by a human to comfortably interact with other human or with an object.
\item The score should decrease as the number of people it is interrupting increases.
\item In small rooms with a high number of people, closer distances are acceptable in comparison to big rooms with fewer people. It is somewhat acceptable to get closer to people in crowded environments. Therefore, in general terms, the higher the density, the higher the score.
\item You should consider only social aspects, not the robot's intelligence. Even if the robot seems to be having a close look at one of the walls, it should have a decent score as long as it is not disturbing anyone. The variable to assess is not related to the robot's performance or whether or not the robot collides with walls and~objects. We are only asking about social aspects.
\end{itemize}

\section{Analysis and validation of the dataset}\label{analysis}
\begin{figure}
\centering
    \subcaptionbox{Histogram depicting the categories used by three subjects for 500 scenarios.\label{fig:labelling_data_a}}{\includegraphics[width=.9\textwidth]{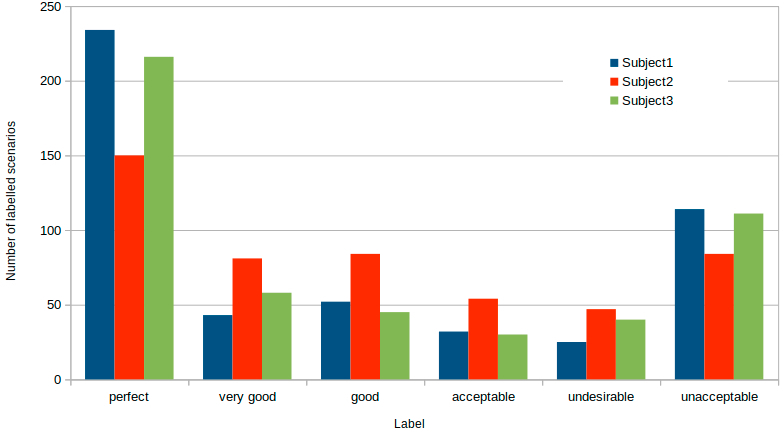}}
    \vfill
    \par\bigskip
    \subcaptionbox{Variation in the labelling of the subjects with respect to the mean for 150 scenarios. The~letters at the top (\textbf{A}--\textbf{D}) refer to the scenarios: \ref{fig:scenario_a}, \ref{fig:scenario_b}, \ref{fig:scenario_c} and~\ref{fig:scenario_d} respectively \label{fig:labelling_data_b}.}{\includegraphics[width=.9\textwidth]{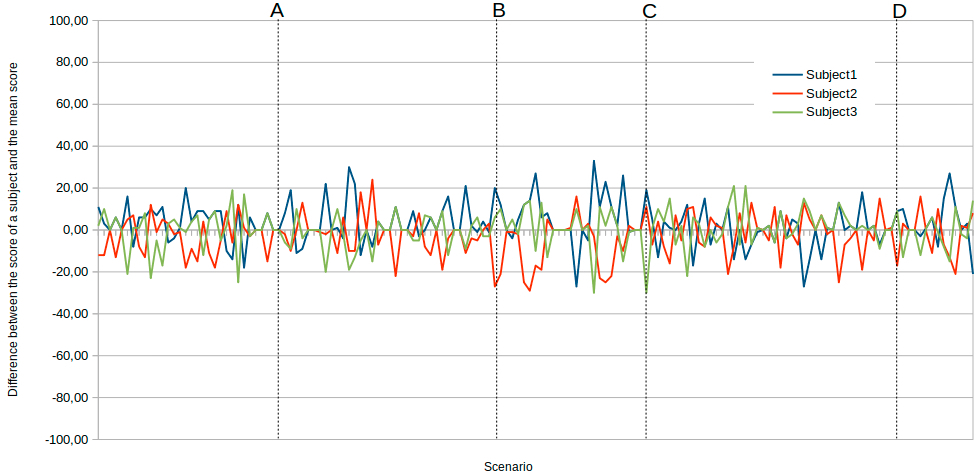}}
    \par\bigskip
    \caption{Overview of the data provided by three different subjects.}\label{fig:labelling_data}
\end{figure}

This section provides a brief analysis of the data to facilitate understanding its relatively subjective nature and~how the labels are distributed.
To this end, a subset of the scenarios which were labelled by~three subjects is used (see Figure \ref{fig:labelling_data}).

\par
Figure \ref{fig:labelling_data_a} depicts a histogram of the labels provided by three different subjects for 500 common scenarios. Each label represents a score range: $[84-100]$ for perfect, $[67-83]$ for very good, $[51-66]$ for good, $[34-50]$ for acceptable, $[17-33]$ for undesirable, and~$[0-16]$ for unacceptable.
From this figure, certain variability on the opinion of the 3 subjects can be observed.
Thus, subjects 1 and~3 tend to give a more extreme score to the different scenarios than subject 2, who scores a higher number of situations with intermediate labels. 
Despite the observed variations, the three subjects assign a score greater than 50, the limit between good and acceptable, to a similar number of scenarios (around 320).
This indicates that~no relevant divergences are found among the different opinions of a common scenario.
\par
In order to quantitatively evaluate the consistency of the labelling of the three users, additional reliability measures were obtained.
In particular, the inter-rater and intra-rater agreement has been computed using the linearly weighted kappa coefficient~\cite{Cohen1968}.
The intra-rater measures have been obtained using a set of 200 scenarios labelled twice by each subject. The inter-rater agreement has been computed from 522 common scenarios labelled by the three subjects.
The results of these measures can be observed in Table \ref{tab:inter_intra}.
The diagonal cells of the Table \ref{tab:inter_intra} represent the intra-rater reliability, while the remaining cells show the inter-rater agreement for every pair of subjects.
As can be seen, the inter-rater consistency ranges from $0.61$ to $0.71$, showing substantial agreement among the three users according to~\cite{Landis77}.
In addition, intra-rater reliability for all the subjects are above $0.75$, which indicates substantial to almost perfect agreement, \textit{i.e.} high reliability.

\begin{table}[ht]
\caption{Inter-rater and intra-rater consistency of three subjects.}
\centering
\begin{tabular}{|c|c|c|c|}
\hline
\textbf{\hfill}	& \textbf{Subject1} & \textbf{Subject2} & \textbf{Subject3}  \\
\hline
\textbf{Subject1} & 0.84 & 0.61 & 0.71\\
\hline
\textbf{Subject2} & 0.61 & 0.76 & 0.71\\
\hline
\textbf{Subject3} & 0.71 & 0.71 & 0.81 \\
\hline
\end{tabular}
\label{tab:inter_intra}
\end{table}

\par
Figure \ref{fig:labelling_data_b} provides additional data that reinforce the above observation.
This figure represents the difference between the score of a subject and~the mean score of the three subjects for 150 common scenarios.
The four scenarios of Figure~\ref{fig:sndg} have been marked in the chart with vertical dotted lines.
The~standard deviation for each of the three subjects considering the 150 scenarios is around 10 points.
Given that this value is lower than the width of the label ranges, the variation of the score provided by~the three subjects can be considered moderately low.
For a considerable number of scenarios, such~as the scenarios in Fig.~\ref{fig:scenario_a} and Fig.~\ref{fig:scenario_d}, the three subjects assign similar scores.
Nevertheless, other scenarios produce more variability.
This is the case of the scenarios in Fig.~\ref{fig:scenario_b} and Fig.~\ref{fig:scenario_c}, which are more susceptible to generate different feelings than those in Fig.~\ref{fig:scenario_a} and Fig.~\ref{fig:scenario_d}.
\par
Considering the whole set of 5735 different scenarios labelled by all the subjects, similar results are observed.
The variability in these scenarios has been individually measured for different subsets, grouping the scenarios according to the number of times they have been labelled. 
For each subset, the pooled standard deviation ($s_p$) (\cite{Cohen1988}) has been computed as a measure of dispersion. 
The pooled standard deviation is a weighted average of standard deviations of several groups.
For each subset, since every scenario is labelled by the same number of subjects, the value of $s_p$ can be computed as follows:

\begin{equation}
    s_p = \sqrt{\frac{\sum_{i=1}^{n} s_i^2}{n}}
\end{equation}

\noindent where $n$ is the number of scenarios of the corresponding subset and $s_i$ the standard deviation of a given scenario.

Using the above measure, the resulting values show that the dispersion remains below $11.5$ points in all the subsets.
Taking into account that the highest difference between two labels is $100$, the obtained values of dispersion can be considered certainly low.
Moreover, combining the dispersion of all the subsets, a global $s_p$ of $9.28$ is obtained, which is in line with the results of Figure~\ref{fig:labelling_data}.
\par
\begin{figure}
\centering
    \subcaptionbox{Simple scenario with two interacting humans.}{\includegraphics[width=.75\textwidth]{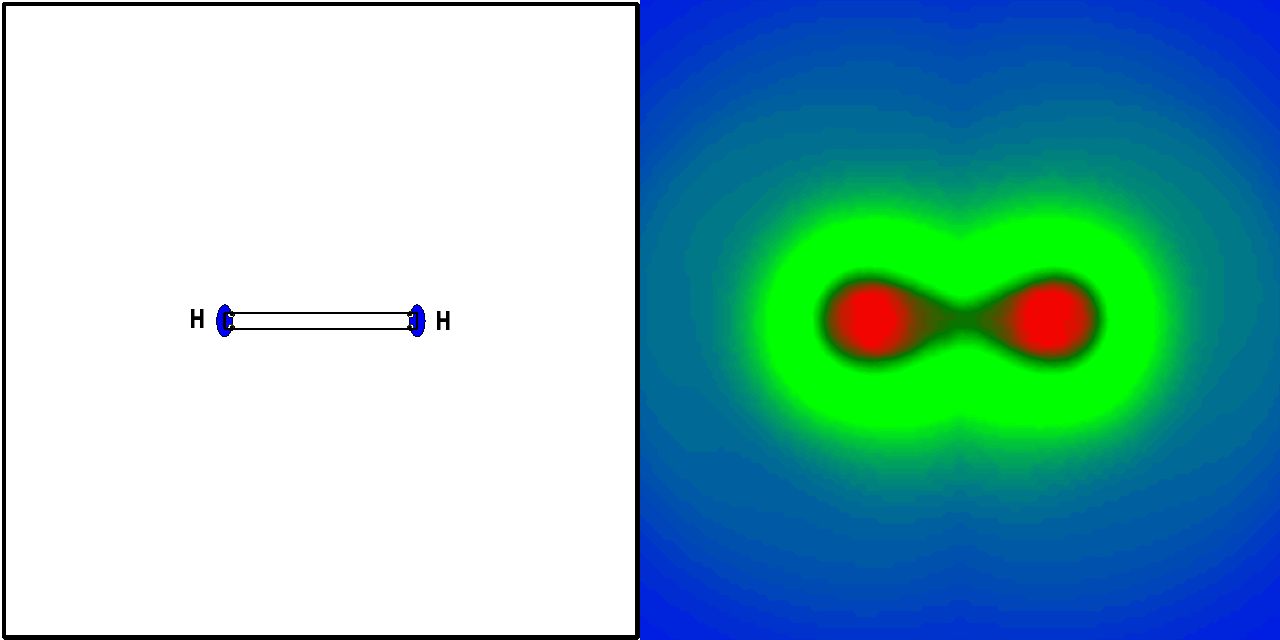}}
    \par\bigskip
    \subcaptionbox{Complex scenario with eight humans and different interactions.}{\includegraphics[width=.75\textwidth]{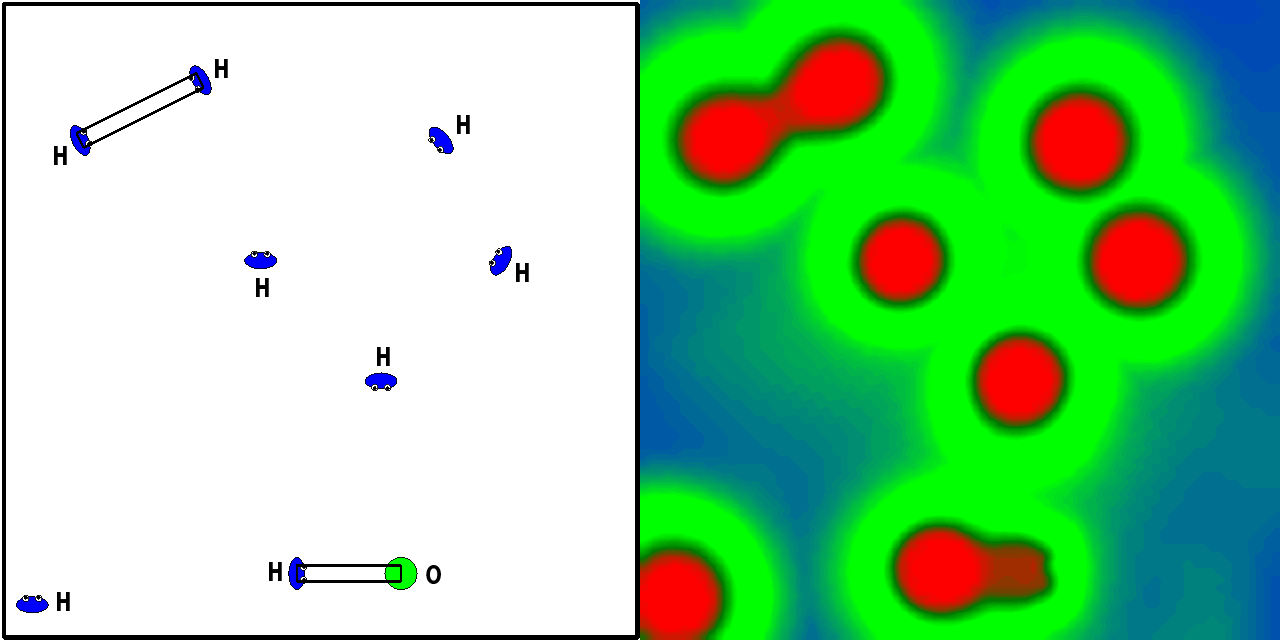}}
    \par\bigskip    
    \caption{Results obtained using a GNN trained with the proposed dataset. The images of the left show different scenarios where different situations occur. The images of the right represent the output of the network for the different positions of the robot in the scene. The response of the network is depicted with a heat colour scale that varies between red (unacceptable position) to blue (perfect position). }\label{fig:gnn_results}
\end{figure}

Besides this analysis, the dataset has been successfully applied to train Graph Neural Networks (GNNs) that model adherence to social-navigation conventions for robots.
Some results are presented in Figure~\ref{fig:gnn_results}.
The figure shows several scenarios (left) and the response of the network for the different locations of the robot in each scenario (right).
The disturbance of the robot in each position of the scene is represented using a heat colour scale, where red indicates unacceptable position and blue expresses perfect position. 
More details about the application of the dataset to GNNs can be found in~\cite{manso2019graph}.

\section{Discussion}\label{discussion}
Datasets are extremely important in many scientific disciplines. They are essential for~benchmarking and~algorithm comparison, but with the emergence of deep learning, datasets have become the basic support over which the new artificial intelligence is sustained.
For the problem at~hand, the variable size of the data and~its structured nature is one of the challenges from a learning point of view.
The 9280 samples generated in SocNav1 seem to be enough for machine learning purposes given the size of the data structures describing the scenarios.
Initial results applying the dataset to graph neural networks support this idea~\cite{manso2019graph}. 
\par
Regarding the design of the experiments, it is worth noting that the labels describe how humans think they would feel in the situation, not how they would feel if they were actually there.
Generating a dataset providing direct measurements would be extremely challenging from a technological point of~view (how are the measurements taken) as well as from a managerial perspective (time and~resources needed). Nevertheless, the main reason for constructing a synthetic dataset is based on the very nature of the data. The aim is to include scenarios that offer the greatest range of possibilities. If limited by~real scenarios, situations that could endanger the integrity of the human being, such as, for example, those that represent a robot-human collision, could not have been included.
\par
Even though the data for each scenario might not seem very complex, as mentioned in the~introduction in more detail, the datasets currently available do not consider interactions between people or objects.
We are however open to extending the dataset with new features in the future if~it~is~found useful.


\vspace{6pt} 

\section*{Author Contributions}
All authors contributed to the design of the dataset and~the manuscript.
All authors have read and agreed to the published version of the manuscript.
The software to collect and~process the data was developed~by~L.J.M. and~P.B.

\section*{Funding}
This work has partly been supported by grants 0043\_EUROAGE\_4\_E, from the European Union ---Interreg project--- and~by grants GR18133 and~IB18056, from the Government of Extremadura. It also has been partially funded by the Institute of Coding, a UK national initiative supported by the government via a £20-million grant from the Office for Students.


\section*{Conflicts of Interest}
The authors declare that the research was conducted in the absence of any commercial or~financial relationships that could be construed as a potential conflict of interest.

\section*{Data Availability Statement}
The dataset compiled can be found in the GitHub repository \textit{ljmanso/SocNav1}: \url{https://github.com/ljmanso/SocNav1}.

 \bibliography{preprint}{}

\begin{thebibliography}{10}

\bibitem{Battaglia2018}
Peter Battaglia, Jessica Blake~Chandler Hamrick, Victor Bapst, Alvaro Sanchez,
  Vinicius Zambaldi, Mateusz Malinowski, Andrea Tacchetti, David Raposo, Adam
  Santoro, Ryan Faulkner, Caglar Gulcehre, Francis Song, Andy Ballard, Justin
  Gilmer, George~E. Dahl, Ashish Vaswani, Kelsey Allen, Charles Nash,
  Victoria~Jayne Langston, Chris Dyer, Nicolas Heess, Daan Wierstra, Pushmeet
  Kohli, Matt Botvinick, Oriol Vinyals, Yujia Li, and Razvan Pascanu.
\newblock Relational inductive biases, deep learning, and graph networks.
\newblock {\em arXiv}, 2018.

\bibitem{Borkowski2010}
Adam Borkowski, Barbara Siemiatkowska, and Jacek Szklarski.
\newblock {Towards semantic navigation in mobile robotics}.
\newblock {\em Lecture Notes in Computer Science (including subseries Lecture
  Notes in Artificial Intelligence and Lecture Notes in Bioinformatics)}, 5765
  LNCS:719--748, 2010.

\bibitem{Cohen1968}
Jacob Cohen.
\newblock Weighted kappa: Nominal scale agreement provision for scaled
  disagreement or partial credit.
\newblock {\em Psychological Bulletin}, 70(4):213, 1968.

\bibitem{Cohen1988}
Jacob Cohen.
\newblock {\em Statistical power ANALYSIS for the Behavioral sciences}, volume
  2nd.
\newblock Lawrence Erlbaum Associates, 01 1988.

\bibitem{Cruz-maya2019}
Arturo Cruz-Maya, Fernando Garcia, and Amit~K. Pandey.
\newblock {Enabling Socially Competent navigation through incorporating HRI}.
\newblock {\em arXiv}, pages 9--12, 2019.

\bibitem{Fisher2004}
Robert~B. Fisher.
\newblock {The PETS04 surveillance ground-truth data sets}.
\newblock Technical report, School of Informatics, University of Edinburgh, 1
  2004.

\bibitem{fosch2018did}
Eduard Fosch~Villaronga, Aurelia Tam{\`o}-Larrieux, and Christoph Lutz.
\newblock {Did I Tell You My New Therapist is a Robot? Ethical, Legal, and
  Societal Issues of Healthcare and Therapeutic Robots}.
\newblock {\em Ethical, Legal, and Societal Issues of Healthcare and
  Therapeutic Robots (October 17, 2018)}, 2018.

\bibitem{garcia2018challenges}
Angel Garc{\'\i}a-Olaya, Raquel Fuentetaja, Javier Garc{\'\i}a-Polo,
  Jos{\'e}~Carlos Gonz{\'a}lez, and Fernando Fern{\'a}ndez.
\newblock {Challenges on the Application of Automated Planning for
  Comprehensive Geriatric Assessment Using an Autonomous Social Robot}.
\newblock In {\em Workshop of Physical Agents}, pages 179--194. Springer, 2018.

\bibitem{hall1969hidden}
Edward~T Hall.
\newblock The hidden dimension: man's use of space in public and private the
  bodley head.
\newblock {\em London, Sydney, Toronto}, 121, 1969.

\bibitem{hansen2009adaptive}
Soren~Tranberg Hansen, Mikael Svenstrup, Hans~Jorgen Andersen, and Thomas Bak.
\newblock Adaptive human aware navigation based on motion pattern analysis.
\newblock In {\em RO-MAN 2009-The 18th IEEE International Symposium on Robot
  and Human Interactive Communication}, pages 927--932. IEEE, 2009.

\bibitem{Landis77}
J.~Richard Landis and Gary~G. Koch.
\newblock The measurement of observer agreement for categorical data.
\newblock {\em Biometrics}, 33(1), 1977.

\bibitem{Luber2012}
Matthias Luber, Luciano Spinello, Jens Silva, and O.~Kai Arras.
\newblock {Socially-Aware Robot Navigation: A Learning Approach}.
\newblock In {\em IEEE/RSJ International Conference on Intelligent Robots and
  Systems}, pages 902--907. IEEE, 2012.

\bibitem{manso2019graph}
Luis~J. Manso, Ronit~R. Jorvekar, Diego~R. Faria, Pablo Bustos, and Pilar
  Bachiller.
\newblock Graph neural networks for human-aware social navigation, 2019.

\bibitem{Papadakis2013}
Panagiotis Papadakis, Anne Spalanzani, and Christian Laugier.
\newblock {Social Mapping of Human-Populated Environments by Implicit Function
  Learning}.
\newblock In {\em IEEE/RSJ International Conference on Intelligent Robots and
  Systems}, pages 1701--1707. IEEE, 2013.

\bibitem{Pellegrini2009}
S.~Pellegrini, A.~Ess, K.~Schindler, and L.~{Van Gool}.
\newblock {You'll never walk alone: Modeling social behavior for multi-target
  tracking}.
\newblock In {\em Proceedings of the IEEE International Conference on Computer
  Vision}, pages 261--268, 2009.

\bibitem{perez2018learning}
No{\'e} P{\'e}rez-Higueras, Fernando Caballero, and Luis Merino.
\newblock Learning human-aware path planning with fully convolutional networks.
\newblock In {\em 2018 IEEE International Conference on Robotics and Automation
  (ICRA)}, pages 1--5. IEEE, 2018.

\bibitem{pham2018impact}
Q-C Pham, Raj Madhavan, Ludovic Righetti, William Smart, and Raja Chatila.
\newblock {The Impact of Robotics and Automation on Working Conditions and
  Employment: Ethical, Legal, and Societal Issues}.
\newblock {\em IEEE Robotics \& Automation Magazine}, 25(2):126--128, 2018.

\bibitem{rajnai2017labor}
Zolt{\'a}n Rajnai and Istv{\'a}n Kocsis.
\newblock {Labor market risks of industry 4.0, digitization, robots and AI}.
\newblock In {\em 2017 IEEE 15th International Symposium on Intelligent Systems
  and Informatics (SISY)}, pages 000343--000346. IEEE, 2017.

\bibitem{Ramon-Vigo2014}
Rafael Ramon-Vigo, Noe Perez-Higueras, Fernando Caballero, and Luis Merino.
\newblock {Transferring human navigation behaviors into a robot local planner}.
\newblock In {\em The 23rd IEEE International Symposium on Robot and Human
  Interactive Communication}, pages 774--779. IEEE, 2014.

\bibitem{Rios-Martinez2015}
J.~Rios-Martinez, A.~Spalanzani, and C.~Laugier.
\newblock {From Proxemics Theory to Socially-Aware Navigation: A Survey}.
\newblock {\em International Journal of Social Robotics}, 7(2):137--153, 2015.

\bibitem{Trinh2015}
Thanh~Q. Trinh, Christof Schroeter, Jens Kessler, and Horst-Michael Gross.
\newblock {“Go Ahead, Please”: Recognition and Resolution of Conflict
  Situations in Narrow Passages for Polite Mobile Robot Navigation}.
\newblock In {\em International Conference on Social Robotics}, pages 643--653.
  Springer, 2015.

\bibitem{vega2019socially}
Araceli Vega, Luis~J Manso, Douglas~G Macharet, Pablo Bustos, and Pedro
  N{\'u}{\~n}ez.
\newblock Socially aware robot navigation system in human-populated and
  interactive environments based on an adaptive spatial density function and
  space affordances.
\newblock {\em Pattern Recognition Letters}, 118:72--84, 2019.

\end{thebibliography}
\bibliographystyle{plain}

\end{document}